\documentclass{bmcart}

\usepackage{graphicx}
\usepackage[utf8]{inputenc} %unicode support
\usepackage{amsmath}
\usepackage{epstopdf}
\usepackage{color}
\usepackage{hyperref}%PDF×Ô¶¯±êÇ©

\usepackage{bm}
\usepackage{amssymb}
\usepackage{extarrows}

\startlocaldefs
\endlocaldefs

\begin{document}

%%% Start of article front matter
\begin{frontmatter}

\begin{fmbox}
%\dochead{Research}


\title{Sparse Signal Subspace Decomposition Based on Adaptive Over-complete Dictionary}

%%%%%%%%%%%%%%%%%%%%%%%%%%%%%%%%%%%%%%%%%%%%%%
%%                                          %%
%% Enter the authors here                   %%
%%                                          %%
%% Specify information, if available,       %%
%% in the form:                             %%
%%   <key>={<id1>,<id2>}                    %%
%%   <key>=                                 %%
%% Comment or delete the keys which are     %%
%% not used. Repeat \author command as much %%
%% as required.                             %%
%%                                          %%
%%%%%%%%%%%%%%%%%%%%%%%%%%%%%%%%%%%%%%%%%%%%%%
\author{SUN Hong
\thanks{SUN Hong is with the School of Electronic Information, Wuhan University, Luojia Hill, 430072 Whuhan, China and Signal and Image
Processing Department, Telecom ParisTech, 46 rue Barrault, 75013 Paris, France.}}
\author{SANG Cheng-wei and LE RUYET Didier
\thanks{SANG Cheng-wei is with the School of Electronic Information, Wuhan University, Luojia Hill, 430072 Whuhan, China.}}
\author{LE RUYET Didier
\thanks{LE RUYET Didier is with CEDRIC Laboratory, CNAM,292 rue Saint Martin, 75003 Paris, France.}}

%%%%%%%%%%%%%%%%%%%%%%%%%%%%%%%%%%%%%%%%%%%%%%
%%                                          %%
%% Enter short notes here                   %%
%%                                          %%
%% Short notes will be after addresses      %%
%% on first page.                           %%
%%                                          %%
%%%%%%%%%%%%%%%%%%%%%%%%%%%%%%%%%%%%%%%%%%%%%%

\begin{artnotes}
%\note{Sample of title note}     % note to the article
%\note[id=n1]{Equal contributor} % note, connected to author
\end{artnotes}

\end{fmbox}% comment this for two column layout

%%%%%%%%%%%%%%%%%%%%%%%%%%%%%%%%%%%%%%%%%%%%%%
%%                                          %%
%% The Abstract begins here                 %%
%%                                          %%
%% Please refer to the Instructions for     %%
%% authors on http://www.biomedcentral.com  %%
%% and include the section headings         %%
%% accordingly for your article type.       %%
%%                                          %%
%%%%%%%%%%%%%%%%%%%%%%%%%%%%%%%%%%%%%%%%%%%%%%

\begin{abstractbox}

\begin{abstract} % abstract
%\parttitle{First part title} %if any
%Text for this section.
This paper proposes a subspace decomposition method based on an over-complete dictionary in sparse representation, called "Sparse Signal Subspace Decomposition"
(or 3SD) method. This method makes use of a novel criterion based on
the occurrence frequency of atoms of the dictionary over the data
set. This criterion, well adapted to subspace-decomposition over a
dependent basis set,  adequately reflects the intrinsic
characteristic of regularity of the signal. The 3SD method combines
variance, sparsity and component frequency criteria into an unified
framework. It takes benefits from using an over-complete dictionary
which preserves details and from subspace decomposition which
rejects strong noise. The 3SD method is very simple with a linear
retrieval operation. It does not require any prior knowledge on
distributions or parameters. When applied to image denoising, it
demonstrates high performances both at preserving fine details and
suppressing strong noise.
%\parttitle{Second part title} %if any
%Text for this section.
\end{abstract}

%%%%%%%%%%%%%%%%%%%%%%%%%%%%%%%%%%%%%%%%%%%%%%
%%                                          %%
%% The keywords begin here                  %%
%%                                          %%
%% Put each keyword in separate \kwd{}.     %%
%%                                          %%
%%%%%%%%%%%%%%%%%%%%%%%%%%%%%%%%%%%%%%%%%%%%%%


% MSC classifications codes, if any
%\begin{keyword}[class=AMS]
%\kwd[Primary ]{}
%\kwd{}
%\kwd[; secondary ]{}
%\end{keyword}

\end{abstractbox}
%
%\end{fmbox}% uncomment this for twcolumn layout

\end{frontmatter}

%%%%%%%%%%%%%%%%%%%%%%%%%%%%%%%%%%%%%%%%%%%%%%
%%                                          %%
%% The Main Body begins here                %%
%%                                          %%
%% Please refer to the instructions for     %%
%% authors on:                              %%
%% http://www.biomedcentral.com/info/authors%%
%% and include the section headings         %%
%% accordingly for your article type.       %%
%%                                          %%
%% See the Results and Discussion section   %%
%% for details on how to create sub-sections%%
%%                                          %%
%% use \cite{...} to cite references        %%
%%  \cite{koon} and                         %%
%%  \cite{oreg,khar,zvai,xjon,schn,pond}    %%
%%  \nocite{smith,marg,hunn,advi,koha,mouse}%%
%%                                          %%
%%%%%%%%%%%%%%%%%%%%%%%%%%%%%%%%%%%%%%%%%%%%%%

%%%%%%%%%%%%%%%%%%%%%%%%% start of article main body
% <put your article body there>

%%%%%%%%%%%%%%%%
%% Background %%
%%
\section{INTRODUCTION}

Signal subspace methods (SSM) are efficient techniques to reduce dimensionality of data and to filter out noise \cite{one}.
The fundamental idea under SSM is to project the data on a basis made of two subspaces, one mostly containing the signal and the other the noise.
 The two subspaces are separated by a thresholding criterion associated with some measures of information.

The two most popular methods of signal subspace decomposition are
wavelet shrinkage \cite{two} and Principal Component Analysis
(PCA) \cite{three}. Both techniques have proved to be quite
efficient. However, wavelet decomposition depending on signal
statistics is not equally adapted to different data,
 and requires some knowledge on prior distributions or parameters of signals to efficiently choose the thresholds for shrinkage.
 A significant advantage of the PCA is its adaptability  to data. The separation criterion is based on energy which may be seen as a limitation in some cases as illustrated in the next section.

In recent years, sparse coding has attracted significant interest in
the field of signal denoising \cite{four}. A sparse representation
is a decomposition of a signal on a very small set of components of
an over-complete basis (called dictionary) which is adapted to the
processed data. A difficult aspect for signal subspace decomposition
based on such a sparse representation is to define the most
appropriate criterion to identify the principal components (called
atoms) from the learned dictionary to build the principal
subspace.  The non-orthogonal property of the dictionary does not
allow to use the energy criterion for this purpose, as done with
PCA.

To solve this problem, we introduce a new criterion to measure the importance of atoms and propose a SSM under the criterion of the occurrence frequency of atoms. We thus make benefit both from the richness of over-complete dictionaries which preserves details of information and from signal subspace decomposition which rejects strong noise.

The remainder of this paper is organized as follows: Section 2
presents two related works to signal decomposition. Section  3
introduce the proposed sparse signal subspace decomposition based on
adaptive over-complete dictionary. Some experimental results and analysis are presented in Section 4. Finally,
we draw the conclusion in section 6.

\section{Review of PCA and Sparse Coding Methods}\label{Sec_2}
We start with a brief description of two well-established approaches to signal decomposition that are relevant and related to the approach proposed in the next section.
\subsection{PCA based Subspace Decomposition}
The basic tool of SSM is principal component analysis (PCA). PCA makes use of an orthonormal basis to capture on a small set of vectors (the signal subspace) as much energy as possible from the observed data. The other basis vectors are expected to contain noise only and the signal projection on these vectors is rejected.

Consider a data set $\{\mathbf{x}_m \in \mathbb{R}^{N \times 1}\}_{m=1}^M$ grouped in a matrix form $\mathbf{X}$ of size $N \times M$: $\mathbf{X} = \{\mathbf{x}_m\}_{m=1}^M$. The PCA is based on
singular value decomposition (SVD) with singular values $\sigma_i$
in descending order obtained from:
 \begin{equation}\label{Eq_1}
 \mathbf{X} = \mathbf{U}\mathbf{A} = \mathbf{U}\mathbf{\Sigma} \mathbf{V}^T
 \end{equation}
where $\mathbf{U}$ and  $\mathbf{V}$ are unitary matrices of size $N
\times N$ and $M \times M$ respectively
($\mathbf{U}^T\mathbf{U}=\mathbf{I}_N,
\mathbf{V}^T\mathbf{V}=\mathbf{I}_M$) and $\mathbf{\Sigma}=
\begin{bmatrix} diag\left[\sigma_1,\cdots,\sigma_r\right],\mathbf{0}
\\ \mathbf{0}
\end{bmatrix}$ of size $N
\times M$ with $\sigma_1 \geq \sigma_2 \geq \cdots \geq
\sigma_r > 0 $, $ \{\sigma_i\}_{i=1}^r $ are positive real known as
the singular values of $\mathbf{X}$ with rank $r$  ($r \leq N$).

Equation (1) can be re-written in a vector form as:
 \begin{equation}\label{Eq_2}
\left[\mathbf{x}_1 \quad \mathbf{x}_2 \cdots \mathbf{x}_m \cdots
\mathbf{x}_M\right] = \left[ \mathbf{u}_1 \quad \mathbf{u}_2 \cdots
\mathbf{u}_n \cdots \mathbf{u}_N \right]. \left[\bm{\alpha}_1 \quad
\bm{\alpha}_2 \cdots \bm{\alpha}_m \cdots \bm{\alpha}_M \right]
 \end{equation}
where $\mathbf{U}=\{\mathbf{u}_n \in \mathbb{R}^{N \times 1}
\}_{n=1}^N$ and $\mathbf{A}=\{ \bm{\alpha}_m \in \mathbb{R}^{N
\times 1} \}_{m=1}^M$. Equation (2)  means that the data
set $\{\mathbf{x}_m \}_{m=1}^M$  is expressed on the orthonormal
basis $\{\mathbf{u}_n \}_{n=1}^N$ as $\{ \bm{\alpha}_m  \}_{m=1}^M$.

In the SVD decomposition given in equation (1), the
standard deviation $\sigma_i$ is used as the measurement for
identifying the meaningful basis vector $\mathbf{u}_i$. PCA takes
the first $P(P < r)$ components $\{\mathbf{u}_n \}_{n=1}^P$ to span
the signal subspace, and the remainders $\{\mathbf{u}_n
\}_{n=P+1}^r$ are considered in a noise subspace orthogonal to the
signal subspace. Therefore, projection on the signal subspace will
hopefully filter out noise and reveal hidden structures. The
reconstructed signal $\hat{\mathbf{S}}_{PCA}$ of size $N \times M$
is obtained by projecting in the signal subspace as
 \begin{equation}
 \begin{aligned}
   \hat{\mathbf{S}}_{PCA}=\left[\mathbf{u}_1 \cdots \mathbf{u}_P \quad \mathbf{0}_{P+1} \cdots \mathbf{0}_N \right] . \left[\bm{\alpha}_1 \quad \bm{\alpha}_2 \cdots \bm{\alpha}_m  \cdots \bm{\alpha}_M \right]
   \end{aligned}
 \end{equation}
The underlying assumption is that information in the data set is
almost completely contained in a small linear subspace of the
overall space of possible data vectors, whereas additive noise is
typically distributed through the larger space isotropically. PCA,
using the standard deviation as a criterion, implies that the
components of the signal of interest in the data set have a maximum
variance and the other components are mainly due to the noise.
However, in many practical cases, some components with low variances
might actually be important because they carry information relative
to the signal details. On the contrary, when dealing with noise with
non-Gaussian statistics, it may happen that some noise components
may actually have higher variances. At last, note that it is often
difficult to provide a physical meaning to the orthonormal basis
$\{\mathbf{u}_i \}_{i=1}^r$ of the SVD decomposition (equation
2) although they have a very clear definition in the
mathematical sense as orthogonal, independent and normal. It is
therefore difficult to impose known constraints on the signal
features when they exist after the principal component
decomposition.

\subsection{Sparse Decomposition}
Recent years have shown a growing interest in research on sparse
decomposition of $M$ observations $\{\mathbf{x}_m \in
\mathbb{R}^N\}_{m=1}^M$ based on a dictionary
$\mathbf{D}=\{\mathbf{d}_k \}_{k=1}^K \in \mathbb{R}^{N \times K}$.
When $K>N$, the dictionary is said over-complete. $\mathbf{d}_k \in
\mathbb{R}^N$ is a basis vector, also called atom since they are not
necessarily independent. By learning from data set $\{\mathbf{x}_m
\}_{m=1}^M$, the sparse decomposition is the solution of equation
(4) \cite{four}:
 \begin{equation}\label{Eq_4}
\begin{aligned}
  \{\mathbf{D},\bm{\alpha}_m\}=\operatorname*{arg min}\limits_{\mathbf{D},{\bm \alpha_m}} \parallel {\bm \alpha_m} \parallel_0 +  \parallel \mathbf{D}{\bm \alpha}_m - \mathbf{x}_m\parallel^2_2 \leq \varepsilon, \quad 1 \leq m \leq M
   \end{aligned}
\end{equation}
where $\bm{\alpha}_m=\left[ \alpha_m(1) \; \alpha_m(2) \; \dots \;
\alpha_m({K}) \right]^T \in \mathbb{R}^{K \times 1}$ is the sparse
code of the observation $\mathbf{x}_m$. The allowed error tolerance
$\varepsilon$ can be chosen according to the standard deviation of
the noise. An estimate of the underlying signal
$\{\mathbf{s}_m\}_{m=1}^M$  embedded in the observed data set
$\{\mathbf{x}_m \}_{m=1}^M$ would be:
 \begin{equation}\label{Eq_5}
\begin{aligned}
 \left[\hat{\mathbf{s}}_1 \quad \hat{\mathbf{s}}_2 \cdots \hat{\mathbf{s}}_m \cdots \hat{\mathbf{s}_M} \right] =
  \left[ \mathbf{d}_1 \quad \mathbf{d}_2 \cdots \mathbf{d}_k \cdots \mathbf{d}_K \right]& .
   \left[\bm{\alpha}_1 \quad \bm{\alpha}_2 \cdots \bm{\alpha}_m \cdots \bm{\alpha}_M \right] &\\ \mbox{or equivalently}
   \quad \hat{\mathbf{S}} = \mathbf{D} \mathbf{A}
   \end{aligned}
\end{equation}
where the matrix $\mathbf{A}$  of size $K \times M$ is composed of
$M$ sparse column vectors $\bm{\alpha}_m$.

The first term on the right side of equation (4) is a
sparsity-inducing regularization that constrains the solution with
the fewest number of nonzero coefficients in each of sparse code
vectors $\bm{\alpha}_m (1 \leq m \leq M)$. The underlying assumption
is that a meaningful signal could be represented by combining few
atoms. This learned dictionary adapted to sparse signal descriptions
has proved to be more effective in signal reconstruction and
classification tasks than PCA method, which is demonstrated in the
next section. The second term in equation (4) is the
residual of the reconstruction, based on the mean-square
reconstruction error estimate in the same way as in PCA method.

On the other hand, we note that the dictionary $\mathbf{D}$, a basis in sparse decomposition, is produced by learning noisy data set $\{\mathbf{x}_m \}_{m=1}^M$, so the basis vectors $\{\mathbf{d}_k \}_{k=1}^K$ should be decomposed into a principal subspace and a residual subspace. However, it is impossible to exploit an energy-constrained subspace since $\{\mathbf{d}_k \}_{k=1}^K$ are not necessarily orthogonal or independent.

\section{The Proposed Sarse Signal Subspace Decomposition (3SD)}\label{Sec_3}
In this section, we introduce a novel criterion to the subspace decomposition over a learned dictionary and a corresponding index of significance of the atoms. Then we propose a signal sparse subspace decomposition (3SD) method under this new criterion.
\subsection{Weight Vectors of Learned Atoms}
At first, we intend to find out the weight of the atoms. In the sparse
representation given in (5), coefficient matrix
$\mathbf{A}$ is composed by $M$ sparse column vectors $\bm{\alpha}_m
$, each $\bm{\alpha}_m $ representing the weight of the observation
$\mathbf{x}_m$, a local parameter for the  $m^{th}$ observation. Let us
consider the row vectors $\{\bm{\beta}_k\}_{k=1}^K$ of coefficient
matrix $\mathbf{A}$ :
 \begin{equation}\label{Eq_6}
\begin{aligned}
 \mathbf{A}& = \left[\bm{\alpha}_1 \; \bm{\alpha}_2 \; \cdots \; \bm{\alpha}_M \right]\\& = \begin{bmatrix} \alpha_1(1) \quad \alpha_2(1) \quad \cdots \quad \alpha_M(1) \\ \alpha_1(2) \quad \alpha_2(2) \quad \cdots \quad \alpha_M(2)\\\vdots \quad \quad \quad \vdots \quad \quad \ddots \quad \quad \vdots \\ \alpha_1(K) \quad \alpha_2(K) \quad \cdots \quad \alpha_M(K)   \end{bmatrix} =  \begin{bmatrix} \bm{\beta}_1 \\ \bm{\beta}_2 \\ \vdots \\ \bm{\beta}_K \end{bmatrix}  \\&
 \mbox{where} \quad \bm{\beta}_k=\left[ \alpha_1(k)  \; \alpha_2(k) \; \dots \; \alpha_M(k) \right] \in \mathbb{R}^{1 \times M}
   \end{aligned}
\end{equation}
Note that the row vector $\bm{\beta}_k$ is not necessarily sparse.
Then equation (5) can be rewritten as:
 \begin{equation}\label{Eq_7}
\begin{aligned}
   \hat{\mathbf{S}}& = \mathbf{D} \mathbf{A} \\& = \left[\mathbf{d}_1 \cdots \mathbf{d}_k \cdots \mathbf{d}_K \right] . \left[\bm{\beta}_1^T \cdots \bm{\beta}_k^T \cdots \bm{\beta}_K^T \right]^T
   \end{aligned}
\end{equation}

Equation (7) means that the row vector $\bm{\beta}_k$  is
the weight of the atom $\mathbf{d}_k$, which is a global parameter
over the data set $\mathbf{X}$. Denoting $\|\bm{\beta}_k\|_0$ the
$\ell^0$ zero pseudo-norm of $\bm{\beta}_k$. $\|\bm{\beta}_k\|_0$ is
the number of occurrences of atom $\mathbf{d}_k$ over the data set
$\{\mathbf{x}_m \}_{m=1}^M$. We call it the frequency of the atom
$\mathbf{d}_k$ denoted by $f_k$:

\begin{equation}\label{Eq_8}
\begin{aligned}
   f_k\triangleq Frequency(\mathbf{d}_k | \mathbf{X}) = \| \bm{\beta}_k \|_0
   \end{aligned}
\end{equation}
In the sparse decomposition, basis vectors $\{\mathbf{d}_k \}_{k=1}^K$ are prototypes of signal segments. That allows us to take them as a signal patterns. Thereupon, some important features of this signal pattern could be considered as a criterion to identify significant atoms. It is demonstrated \cite{five} that $f_k$  is a good description of the signal texture. Intuitively, a signal pattern must occur in meaningful signals with higher frequency even with a lower energy. On the contrary, a noise pattern would hardly be reproduced in observed data even with a higher energy.

It is reasonable to take this frequency $f_k$ as a relevance criterion to decompose the over-complete dictionary into a principal signal subspace and a remained noise subspace. Here, we use the word "subspace", but in fact these two subspaces are not necessary independent.

\subsection{Subspace Decomposition Based on Overcomplete Dictionary}
Taking vectors $\{ \bm{\beta}_k\}_{k=1}^K$, we calculate their $\ell^0$-norms $\{\|\bm{\beta}_k \|_0\}_{k=1}^K$ and rank them in descending order as follows. The index $k$ of vectors $\{ \bm{\beta}_k\}_{k=1}^K$ are belonging to the set $\mathbf{C}=\{1,2,\cdots,k,\cdots,K \}$. A one-to-one index mapping function $\pi$ is defined as:
\begin{equation}\label{Eq_9}
\begin{aligned}
   & \quad \quad \quad \quad \pi(\mathbf{C} \rightarrow \mathbf{C}): k=\pi(\tilde{k}), \quad \quad k,\tilde{k} \in \mathbf{C} \\& s.t. \quad \| \bm{\beta}_{\pi(1)} \|_0 \geq \| \bm{\beta}_{\pi(2)} \|_0 \geq \cdots \geq \| \bm{\beta}_{\pi(\tilde{k})} \|_0 \geq \cdots \geq \| \bm{\beta}_{\pi(K)} \|_0
   \end{aligned}
\end{equation}
By the permutation $\pi$ of the row index $k$ of matrix $\mathbf{A} = \left[\bm{\beta}_1^T \cdots \bm{\beta}_k^T \cdots \bm{\beta}_K^T \right]^T$, the reordered coefficient matrix $\tilde{\mathbf{A}}$ becomes
\begin{equation}\label{Eq_10}
\begin{aligned}
  \tilde{\mathbf{A}}&= \left[ \bm{\beta}_{\pi(1)}^T \quad \bm{\beta}_{\pi(2)}^T \cdots \bm{\beta}_{\pi(k)}^T \cdots \bm{\beta}_{\pi(K)}^T \right]^T
   \end{aligned}
\end{equation}
With corresponding reordered dictionary $\tilde{\mathbf{D}} = \{
\mathbf{d}_{\pi(k)} \}_{k=1}^K$, equation (7) can be
written as:
\begin{equation}\label{Eq_11}
\begin{aligned}
 \hat{\mathbf{S}} &= \tilde{\mathbf{D}} \tilde{\mathbf{A}}  \\&= \left[ \mathbf{d}_{\pi(1)} \cdots \mathbf{d}_{\pi(k)} \cdots \mathbf{d}_{\pi(K)} \right] . \left[\bm{\beta}_{\pi(1)}^T \cdots \bm{\beta}_{\pi(k)}^T \cdots \bm{\beta}_{\pi(K)}^T  \right]^T
     \end{aligned}
\end{equation}
Then, the span of the first $P$ atoms can be taken as a principal
subspace $\mathbf{D}_P^{(\mathbf{S})}$ and the remaining atoms span
a noise subspace $\mathbf{D}_{K-P}^{(N)}$ as:
\begin{equation}\label{Eq_12}
\begin{aligned}
  \mathbf{D}_P^{(\mathbf{S})} = span\{\mathbf{d}_{\pi{(1)}},\mathbf{d}_{\pi{(2)}},\cdots,\mathbf{d}_{\pi{(P)}} \} \qquad   \\ \mathbf{D}_{K-P}^{(N)} = span\{\mathbf{d}_{\pi{(P+1)}},\mathbf{d}_{\pi{(P+2)}},\cdots,\mathbf{d}_{\pi{(K)}} \}
   \end{aligned}
\end{equation}

An estimate $\hat{\mathbf{S}}_P$ of the underlying signal $\mathbf{S}$ embedded
in the observed data set $\mathbf{X}$ can be obtained on the
principal subspace $\mathbf{D}_P^{(\mathbf{S})}$ simply by linear
combination:

\begin{equation}\label{Eq_13}
\begin{aligned}
 \hat{\mathbf{S}}_P &= \mathbf{D}_P^{(\mathbf{S})} . \mathbf{A}_P^{(\mathbf{S})}  \\
  &=  \left[\mathbf{d}_{\pi(1)} \cdots \mathbf{d}_{\pi(k)} \cdots \mathbf{d}_{\pi(P)}  \right] . \left[{\bm{\beta}}_{\pi{(1)}}^T \cdots {\bm{\beta}}_{\pi{(k)}}^T \cdots {\bm{\beta}}_{\pi{(P)}}^T \right]
\end{aligned}
\end{equation}

\subsection{Threshold of Atom's Frequency}
Determining the number $P$ of atoms spanning the signal subspace
$\mathbf{D}_{P}^{(\mathbf{S})}$ is always a hard topic especially
for wide-band signals. Here, $P$ is the threshold of atom's
frequency $f_k$ to distinguish a signal subspace and a noise
subspace. One of the advantages of 3SD is that this threshold $P$
can be easily chosen without any prior parameter.

For a noiseless signal even with some week details, such as the
image example in Fig. 1(a), the atoms' frequencies
$f_{\pi(k)}^{image}$s shown in Fig. 1(d) (in black line)
are almost always high except the zero value. For a signal with
strong noise, such as the example in Fig. 1(b), the
atoms' frequencies $f_{\pi(k)}^{noise}$s shown in Fig.
1(d) (in red line) are almost always equal to $1$
without zero and very few with value $2$ or $3$. It is easy to set a
threshold $P$ of $f_k$ (dotted line in the Fig. 1(d)) to
separate signal's atoms from noise's atoms. By contrast, the index
numbers $\|\bm{\beta}_k\|_2$s under energy criterion shown in Fig.
1(c) for this example are rather puzzle to identify
principal bases.

\begin{figure}[!ht]
  \begin{center}
  \includegraphics[width=0.95 \linewidth]{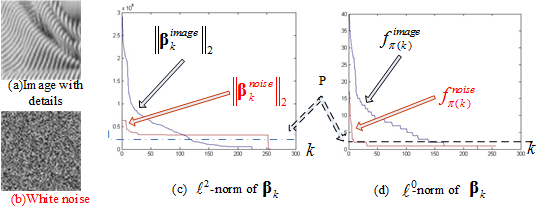}
 \caption{Sparse signal subspaces with criterion of atom's frequency.}\label{figure1}
  \end{center}
\end{figure}

For a noisy signal, such as an image example in Fig.
2(a), its adaptive over-complete dictionary (Fig.
2(b)) consists of atoms of noiseless signal patterns,
pure noise patterns and noisy signal patterns. Signal atoms should
have higher frequencies, noise ones lower and noisy ones moderate.
Intuitively, the red line (Fig. 2(c)) should be a
suitable threshold $P$ of the frequencies $f_k$s. In practical
implementation, the value of $P$ could be simply decided relying to
the histogram of $f_k$. As shown in Fig. 2(d), one can
set the value of $f_k$ associated to the maximum point of its histogram to $P$ as follows:

\begin{equation}\label{Eq_14}
\begin{aligned}
 P =\operatorname*{arg \ max \ Hist}\limits_{k}{(\|\bm{\beta}_k \|_0)}
 \end{aligned}
\end{equation}

\begin{figure}[!ht]
  \begin{center}
  \includegraphics[width=0.8 \linewidth]{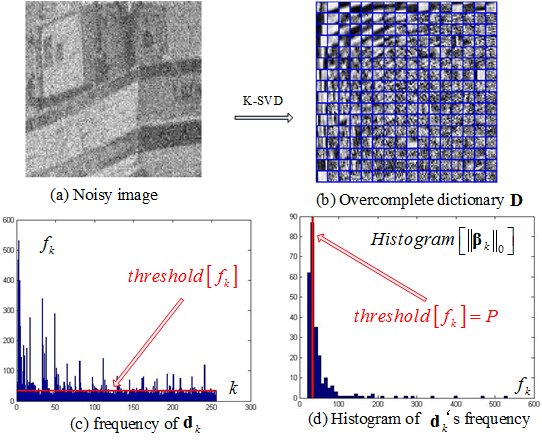}
 \caption{The threshold $P$ of the frequencies $f_k$s.}\label{figure2}
  \end{center}
\end{figure}

In fact, the performances in signal analyses by 3SD method are not
sensitive to the threshold $P$,  owed to the dependence of the
atoms. To demonstrate this point, we take $3$ images, Barbara, Lena and
Boat. Their histograms of $f_k$ are shown in Fig. 3(a)
with the maximum points in dotted lines, $121$, $97$ and $92$
respectively. Fig. 3(b) reports the peak signal-to-noise
ratio (PSNR) of the retrieved images $\hat{\mathbf{S}}_P$ on the
signal principal subspace $\mathbf{D}_{P}^{(\mathbf{S})}$  with
respect of $P$. We can see that PSNRs of the results remain the
same in a large range around the maximum points (in dotted lines).
Consequently, taking the value of $f_k$ associated to the maximum point of its histogram as
the threshold $P$ is a reasonable solution.

\begin{figure}[!ht]
  \begin{center}
  \includegraphics[width=0.95 \linewidth]{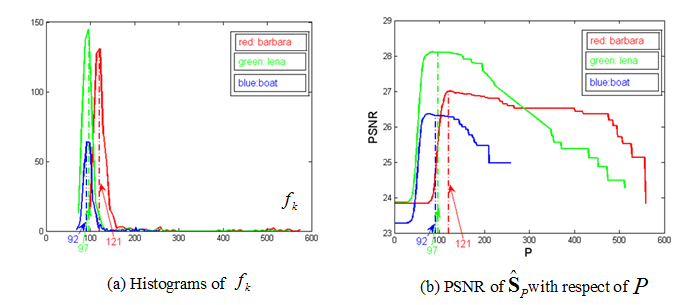}
 \caption{The insensitivity of the threshold $P$.}\label{figure3}
  \end{center}
\end{figure}

\section{Results and Discussion}\label{Sec_4}

\subsection{Signal Decomposition Methods}

Taking a part of the noisy Barbara image (Fig. 4(a)), we
show an example of the sparse signal subspace decomposition (3SD)
and the corresponding retrieved image (Fig. 4(b)). For
comparison, the traditional sparse decomposition and the PCA based
subspace decomposition are shown in Fig. 4(c) and Fig.
4(d).

\begin{figure}[!ht]
  \begin{center}
  \includegraphics[width=0.95 \linewidth]{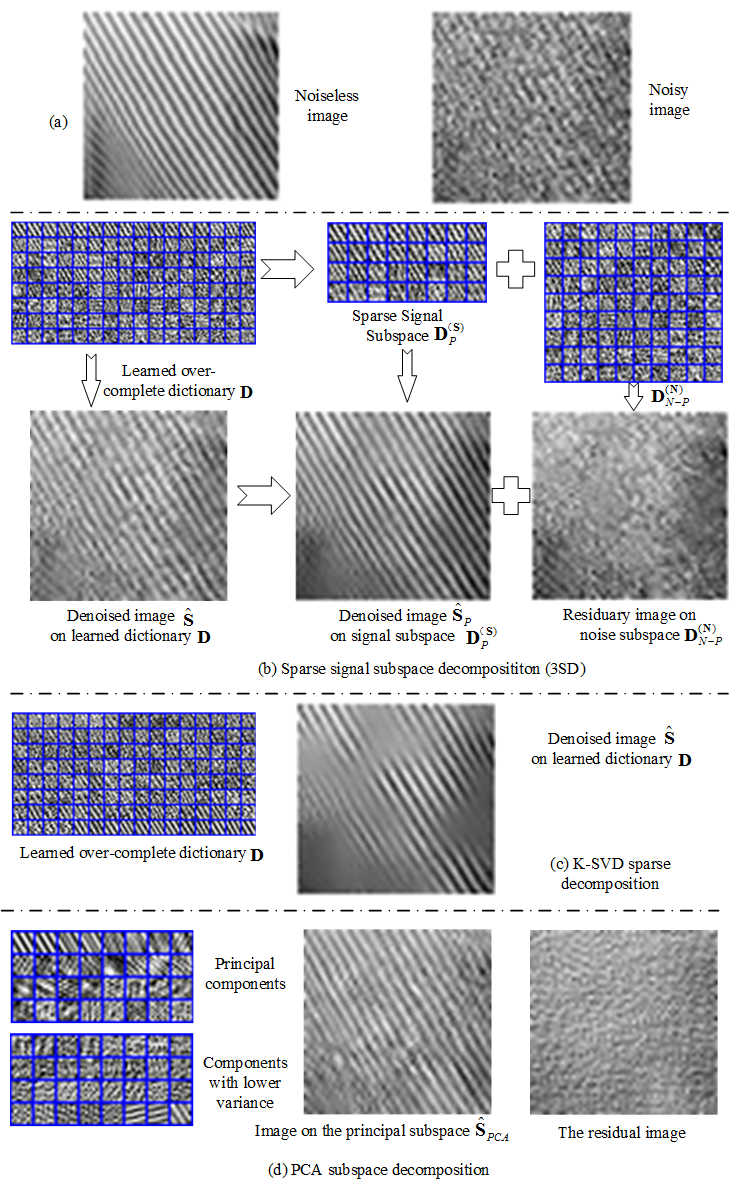}
 \caption{Sparse signal decompositions and principal subspaces.}\label{figure4}
  \end{center}
\end{figure}

Let us look at the proposed sparse signal subspace decomposition on
the top of Fig. 4(b) The $128$ atoms $\mathbf{d}_k$s of
the learned overcomplete dictionary $\mathbf{D}$ are shown in
descending order of their energies measured by $\| \bm{\beta}_k
\|_2$. The $32$ principal signal atoms are chosen from the
dictionary $\mathbf{D}$ under the frequency criterion. They are
shown in descending order of their frequencies measured by $\|
\bm{\beta}_k \|_0$ composing a signal subspace
$\mathbf{D}_{32}^{(\mathbf{S})}$. We can see that some of the
principal atoms are not among the first $32$ atoms with the largest
energy in the overcomplete dictionary $\mathbf{D}$. The retrieved
images are shown on the bottom of Fig. 4(b). The image
$\mathbf{S}$ on $\mathbf{D}$ is apparently denoised. The image
$\hat{\mathbf{S}}$ on the signal subspace
$\mathbf{D}_{32}^{(\mathbf{S})}$ improves obviously by preserving fine details and at suppressing strong noise. On the
other hand, the residual image on noise subspace
$\mathbf{D}_{96}^{(N)}$ contains some very noisy information. This
is because the atoms of the overcomplete dictionary are not
independent.

For the same example, the classical sparse decomposition is shown in
Fig. 4(c). Using K-SVD algorithm \cite{six} in which the
allowed error tolerance $\varepsilon$ (in equation (4)) is
set to a smaller value to filter out noise. The retrieved image
$\mathbf{S}$ seems to have a high Signal to Noise Ratio (SNR), but
have lost the weak information. This is because signal distortion
and residual noise cannot be minimized simultaneously at dictionary
learning by equation (4).

In another comparison, the PCA based subspace decomposition is shown
in Fig. 4(d). The $64$ components are orthonormal and
the $32$ principal components are of the largest variance. The
retrieved image by projecting on the signal subspace is rather
noisy. This is because it cannot suppress strong noise and preserve
weak details of information only using the variance criterion.

\subsection{Application to Image Denoising}

The application of 3SD to image denoising is presented here. A major difficulty of denoising is to separate underlying signal from noise. The proposed 3SD method could win this challenge. In 3SD method, the important components are selected from the over-complete dictionary relying to their occurrence number over the noisy image set. Evidently, the occurrence numbers would be large for signal, even for weak details, such as edges or textures and so on. On the other hand, the occurrence numbers would be low for different kinds of white Gaussian or non-Gaussian noises, even strong at intensity.

The 3SD algorithm for image denoising is presented as follows:

%\begin{table}[h!]
%\caption{ALGORITHM OF IMAGE DENOISING BY 3SD METHOD}
      \begin{tabular}{@{}l @{} l @{}}
        \hline
$\mathbf{Input}$: &\ Noisy image $\mathbf{X}$  \qquad \qquad  \qquad \qquad \qquad \qquad  \qquad \qquad\\
 $\mathbf{Output}$:&\ Denoised image $\hat{\mathbf{S}}$  \\
  \qquad -  &\ Sparse representation $\{\mathbf{D}, \mathbf{A}\}$: using K-SVD algorithm \cite{six} by (4)\\
 \qquad -    &\ Identify principal atoms from $\mathbf{D}$ based on $\mathbf{A}$ :\\
    &\     $\blacksquare$ \qquad Compute the frequencies of atoms $\{\|\bm{\beta}_k \|_0\}_{k=1}^K$ according to (6) and (8)\\
    &\     $\blacksquare$ \qquad Get the permutation $\pi$ sorting the index $k$ of $\{\|\bm{\beta}_k \|_0\}_{k=1}^K$ by (9) \\
    &\     $\blacksquare$ \qquad Compute the threshold $P$ by (14) \\
   \qquad  -  &\     Obtain the signal principal atoms $\{\mathbf{d}_{\pi(k)}\}_{k=1}^P$ by (12)\\
 \qquad  -  &\  Reconstruct image $\hat{\mathbf{S}_P}$ by (13)\\
  \hline
      %  A3 & ..  & .   & .\\ \hline
      \end{tabular}
%\end{table}

In this application, we intend to preserve faint signal details under a situation of strong noise. We use the peak signal-to-noise ratio (PSNR) to assess the noise removal performance:
 \begin{equation}\label{Eq_15}
\begin{aligned}
PSNR=20 \cdot \log_{10} \left[MAX\{\mathbf{S}(i,j) \}\right] -10
\cdot \log_{10} \left[ MSE \right] \\
MSE=\frac{1}{IJ}{\sum\nolimits_{i=0}^{I-1}}{\sum\nolimits_{j=0}^{J-1}}\left[\mathbf{S}(i,j)-\hat{\mathbf{S}}(i,j)
\right]^2
 \end{aligned}
\end{equation}
and the structural similarity index metric (SSIM) between the denoised image   and the pure one   to evaluate the preserving details performance:
 \begin{equation}\label{Eq_16}
\begin{aligned}
SSIM(\mathbf{S},\hat{\mathbf{S}})=\frac{(2u_{\mathbf{S}}u_{\hat{{\mathbf{S}}}} + c_1)(2\sigma_{\mathbf{S}\hat{{\mathbf{S}}}}+c_2) }{(u^2_{\mathbf{S}}+ u^2_{\hat{\mathbf{S}}} +c_1)(\sigma^2_{\mathbf{S}} + \sigma^2_{\hat{\mathbf{S}}}+ c_2) }
 \end{aligned}
\end{equation}
where $u_x$ is the average of $x$, $\sigma_x^2$ is the variance of $x$, $\sigma_{xy}$ is the covariance of $x$ and $y$, and $c_1$ and $c_2$ are small variables to stabilize the division with weak denominator.

In the experiments, dictionaries used $D$s are of size $64 \times 256$ ($K=256$ atoms), designed to handle image patches $\mathbf{x}_m$ of size $N=64=8 \times 8$ pixels.

\subsection{Image Denoising}

A noisy Lena image $\mathbf{X}=\mathbf{S} + \mathbf{V}$ with an additive zero-mean white Gaussian noise $\mathbf{V}$ is used. The standard deviation of noise is $\sigma=35$. A comparison is made with 3SD method and K-SVD method \cite{six} which is one of the best denoising methods reported in the recent literatures.

\begin{figure}[!ht]
  \begin{center}
  \includegraphics[width=0.95 \linewidth]{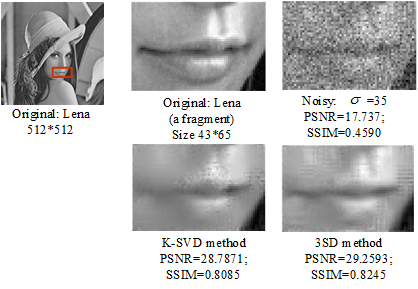}
 \caption{Image denoising comparing the proposed 3SD method with the K-SVD method.}\label{figure5}

  \end{center}
\end{figure}

From the results shown in Fig. 5, the 3SD method
outperforms the K-SVD method by about $1dB$ in PSNR and by about
$1\%$ in SSIM (depending on how much details in images and how faint
the details). In terms of subjective visual quality, we can see that
the corner of mouth and the nasolabial fold with weak intensities
are much better recovered by 3SD method.

\subsection{SAR Image Despeckling}
In the second experiment, a simulated SAR image with speckle noise is used. Speckle is often modeled as multiplicative noise as $x(i,j)=s(i,j)v(i,j)$ where $x$, $s$ and $v$ correspond to the contaminated intensity, the original intensity, and the noise level, respectively.

\begin{figure}[!ht]
  \begin{center}
  \includegraphics[width=0.95 \linewidth]{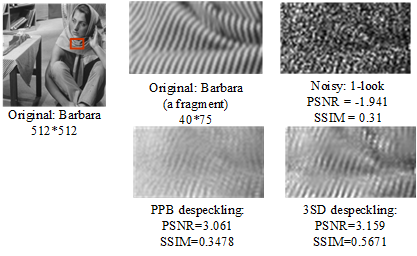}
\caption{SAR image despeckling comparing the proposed 3SD method with the PPB method.}\label{figure6}
  \end{center}
\end{figure}

Fig. 6 shows the despeckling results of simulated
one-look SAR scenario with a fragment of Barbara image. A comparison
is made with 3SD method and a probabilistic patch based (PPB) filter
based on nonlocal means approach \cite{seven} which can cope with
non-Gaussian noise. We can see that PPB can well remove speckle
noise. However, it also removes the low-intensity details. The 3SD
method shows advantages at preserving fine details and at
suppressing strong noise.

\section{Conclusion}\label{Sec_6}
We proposed a method of sparse signal subspace decomposition (3SD).
The central idea of the proposed 3SD is to identify principal atoms
from an adaptive overcomplete dictionary relying the occurrence
frequency of atoms over the data set (equation (8)). The
atoms frequency is measured by zero pseudo-norms of weight
vectors of atoms (equation (6) and
(8)). The principal subspace is spanned by the maximum
frequency atoms (equation (12)).

The 3SD method combines the variance criterion, the sparsity criterion and the component's frequency criterion into a uniform framework. As a result, it can identify more effectively the principal atoms with the three important signal features. On the contrary, PCA uses only variance criterion and sparse coding method uses the variance and the sparsity criterions. In those ways, it is more difficult to distinguish weak information from strong noise.

Another interesting assert of 3SD method is that it takes benefits from using an over-complete dictionary which reserves details of information and from subspace decomposition which rejects strong noise. On the contrary, some undercomplete dictionary methods \cite{eight} and some sparse shrinkage methods \cite{nine,ten} might lose week information when suppressing noise.

Moreover, the 3SD method is very simple with a linear retrieval
operation (equation (13)). It does not require any prior
knowledge on distribution or parameter to determine a threshold
(equation  (14)). On the contrary, some sparse shrinkage
methods, such as \cite{nine}, necessitate non-linear processing with
some prior distributions of signals.

The proposed 3SD could be interpreted as a PCA in sparse decomposition, so it admits straightforward extension to applications of feature extraction, inverse problems, or machine learning.

%\nocite{oreg,schn,pond,smith,marg,hunn,advi,koha,mouse}

%%%%%%%%%%%%%%%%%%%%%%%%%%%%%%%%%%%%%%%%%%%%%%
%%                                          %%
%% Backmatter begins here                   %%
%%                                          %%
%%%%%%%%%%%%%%%%%%%%%%%%%%%%%%%%%%%%%%%%%%%%%%

\begin{backmatter}

\section*{Competing interests}
  The authors declare that they have no competing interests.

\section*{Author's contributions}
    HS proposed the idea, designed the algorithms and experiments, and drafted the manuscript. CWS gave suggestion on the design the algorithm, realized the algorithms, carried out the experiments. DLR gave suggestions on the mathematical expressions of munuscript and experiment analysis as well as explanation, and helped draft the manuscript.

\section*{Acknowledgements}
    This work was supported by the National Natural Science Foundation of China (Grant No. 60872131).
    The idea of the Sparse Signal Subspace decomposition here arises through a lot of deep discussions with Professor Henri MAITRE at Telecom-ParisTech in France, he also gave suggestion on the structure of manuscript.
\end{backmatter}

\begin{thebibliography}{10}

\bibitem{one}
Hermus, K., Wambacq, P., Hamme, H.V.: A review of signal subspace speech enhancement and its application to noise robust speech recognition. EURASIP Journal on Advances in Signal Processing, 888-896 (2007).doi:10.1155/2007/45821
\bibitem{two}
Donoho, D.L., Johnstone, I.M., Kerkyacharian, G., Picard, D.: Wavelet shrinkage: asymptopia? Journal of the Royal Statistical Society. Series B 57, 301-369 (1995). http://citeseer.ist.psu.edu/viewdoc/summary?doi=10.1.1.162.1643
\bibitem{three}
Tufts, D.W., Kumaresan, R., Kirsteins, I.: Data adaptive signal estimation by singular value decomposition of a data matrix. Proceeding of the IEEE 70(6), 684-685 (1982). doi:10.1109/PROC.1982.12367
\bibitem{four}
Elad, M., Aharon, M.: Image denoising via sparse and redundant representations over learned dictionaries. IEEE Trans. Image Processing 15(12), 3736-3745 (2006). doi:10.1109/TIP.2006.881969
\bibitem{five}
Variational texture synthesis with sparsity and spectrum constraints. Journal of Mathematical Imaging and Vision 52, 124-144 (2015). doi:10.1007/s10851-014-0547-7
\bibitem{six}
Aharon, M., Elad, M., Bruckstein, A.: K-svd: an algorithm for designing overcomplete dictionaries for sparse representation. IEEE Trans. Signal Processing 54(11), 4311-4322 (2006). doi:10.1109/TSP.2006.881199
\bibitem{seven}
Deledalle, C.A., Denis, L., Tupin, F.: Iterative weighted maximum likelihood denoising with probabilistic patch-based weights. IEEE Trans. Image Processing 18(12), 2661-2672 (2009). doi:10.1109/TIP.2009.2029593
\bibitem{eight}
Porikli, F., Sundaresan, R., Suwa, K.: Sar depeckling by sparse reconstruction on affinity nets. EUSAR 2012 18,796-799 (2012).
\bibitem{nine}
Hyvarinen, A., Hoyer, P., Oja, E.: Sparse code shrinkage for image denoising. IEEE World Congress on Computational Intelligence 2, 859-864 (1998). doi:10.1109/IJCNN.1998.685880
\bibitem{ten}
Malutan, R., Terebes, R., Germain, C.: Speckle noise removal in ultrasound images using sparse code shrinkage.5th IEEE Conference on E-Health and Bioengineering Conference 2, 1-4 (2015).doi:10.1109/EHB.2015.7391394
\end{thebibliography}
\end{document}